\documentclass[a4paper]{article}

\usepackage{INTERSPEECH2022}
\usepackage{cite}
\usepackage{comment}

\title{Silence is Sweeter Than Speech: \\ Self-Supervised Model Using Silence to Store Speaker Information}
\name{Chi-Luen Feng, Po-chun Hsu, Hung-yi Lee}
\address{
  National Taiwan University
  }
\email{r10944013@ntu.edu.tw, f07942095@ntu.edu.tw, hungyilee@ntu.edu.tw}

\begin{document}

\maketitle
\begin{abstract}

Self-Supervised Learning (SSL) has made great strides recently. SSL speech models achieve decent performance on a wide range of downstream tasks, suggesting that they extract different aspects of information from speech. However, how SSL models store various information in hidden representations without interfering is still poorly understood. Taking the recently successful SSL model, HuBERT, as an example, we explore how the SSL model processes and stores speaker information in the representation. We found that HuBERT stores speaker information in representations whose positions correspond to silences in a waveform. There are several pieces of evidence. (1) We find that the utterances with more silent parts in the waveforms have better Speaker Identification (SID) accuracy. (2) If we use the whole utterances for SID, the silence part always contributes more to the SID task. (3) If we only use the representation of a part of the utterance for SID, the silenced part has higher accuracy than the other parts. Our findings not only contribute to a better understanding of SSL models but also improve performance. By simply adding silence to the original waveform, HuBERT improved its accuracy on SID by nearly 2\%.

\end{abstract}

\section{INTRODUCTION}

Self-Supervised Learning (SSL), which utilizes unlabeled data to learn, has achieved many milestones in different fields. In the speech field, there are several outstanding SSL models proposed in recent years \cite{HuBERT,wav2vec2,vq-wav2vec,wav2vec, mockingjay,decoar2,CPC,APC,vq-apc, audiobert,tera,decoar,distillhubert,data2vec}.
We can leverage representations from these SSL models with a simple downstream model and perform well in different speech tasks~\cite{s3prl}. This shows that the SSL models have universal characteristics. The representations integrate different information, such as content, semantic, and speaker information, into the low dimension representations.

Despite the achievements of SSL models in various speech tasks, there are only limited research and analysis about SSL models and their representations. All previous works focus mainly on the performance of different layers or types of models~\cite{similarity_analysis, layerwise_analysis, Unispeech, chen2021wavlm}. There is currently no research on how the SSL model stores such complex speech information in representations. In this work, we choose speaker information as the entry point to demystify the SSL model.
Instead of layer-wise analysis like in previous works, we propose to analyze the speaker information through the position of the waveform. The way SSL models store speech information into different parts of an utterance hasn't been studied well. We hypothesize that the SSL model may keep different types of information in different positions, helping the model better disentangle and store various speech information.

We adopt Speaker Identification (SID) task to measure speaker information in the representations. We first find that SSL model tends to store speaker information in the last fragment, which contains more silence. We then further analyze the relationship between the percentage of silence and SID accuracy.
To our surprise, the amount of silence in utterances is highly correlated with accuracy on the SID task, which means that silence has some relationship with speaker information for the SSL model. To get more evidence, we add silence in utterances, split them into fragments, and use only one fragment to train the SID task at a time. We find that the fragment corresponding to the silence part can get the highest score among all fragments. Besides, we can get better accuracy in the SID task by simply adding silence during training. All the evidence points out that silence will help the SSL model store speaker information. To the best of our knowledge, this is the first work using position information to analyze the representation characteristic of the SSL model in speech and explore the storage mechanism of the SSL model.





\begin{figure*}[t]
  \centering
  \includegraphics[scale=1.2 ]{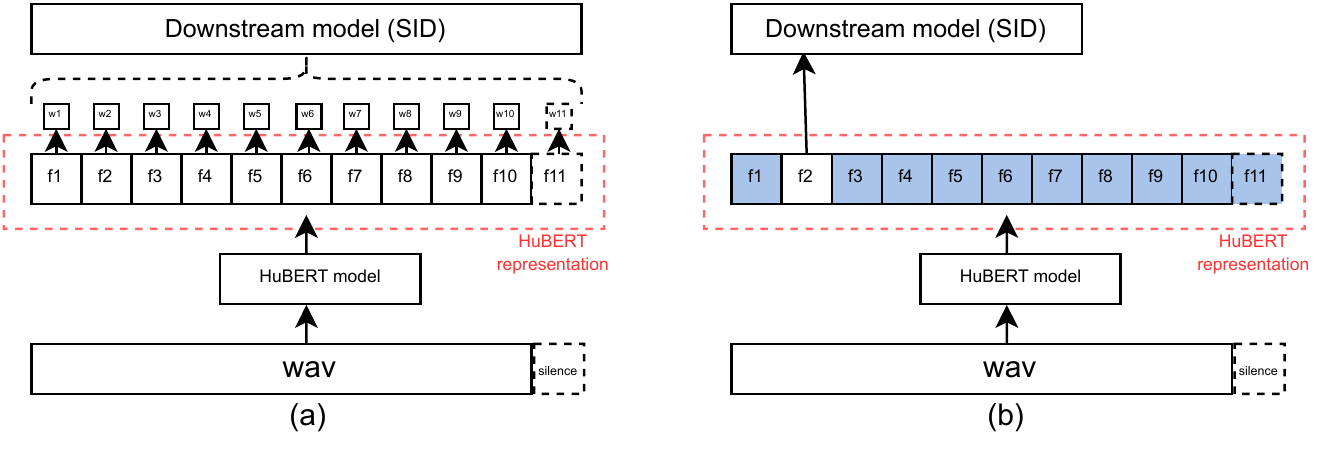}
  \vspace{-10pt}
  \caption{(a) Experiment structure in Section~\ref{sec:WHERE IS THE SPEAKER INFORMATION} and Section~\ref{ssub:silence is important for ssl}. The dotted square represent the silence fragment, which is added only in Section~\ref{ssub:silence is important for ssl}. The input of the downstream model is the weighted sum of the HuBERT representations. 
  (b) Experiment structure in Section~\ref{ssub:do silence really important}. The downstream model takes only one fragment as input at a time.} 
  \vspace{-6pt}
  \label{fig:structure of experiment}
\end{figure*}

\section{RELATED WORK}
According to the success of the SSL model in recent years, there are more and more papers trying to analyze the key to the success of the SSL model. However, the current papers which analyze the SSL model mainly focus on the text field~\cite{ssl_text_1, ssl_text_2, ssl_text_3, ssl_text_4,  ssl_text_5,  ssl_text_6,  ssl_text_7}. For example, \cite{AttentionBert} analyzes how attention maps in BERT process linguistic knowledge. As for~\cite{BERT_RESEARCH}, this paper focuses on how the SSL model learned the language information during the training process. There are still seldom papers contributing to the study of the SSL model in the speech field.

There are two research directions in the limited papers for analysis of the speech SSL model. The first direction is comparing speech SSL models by training criteria. \cite{similarity_analysis} shows that training criteria have a high relationship with the performance of downstream tasks. The second direction is to study the SSL model from the perspective of content. For example, \cite{layerwise_analysis} uses Canonical Correlation Analysis (CCA) to investigate which information is encoded by each layer of the SSL model.
Inspired by these prior works, we want to explore the logic behind the SSL model. However, unlike previous works, we mainly focus on investigating how the information is stored in representations of the SSL model and how the input waveform will cooperate with the SSL model. 

\begin{figure}[t]
  \centering
  \includegraphics[width=\linewidth]{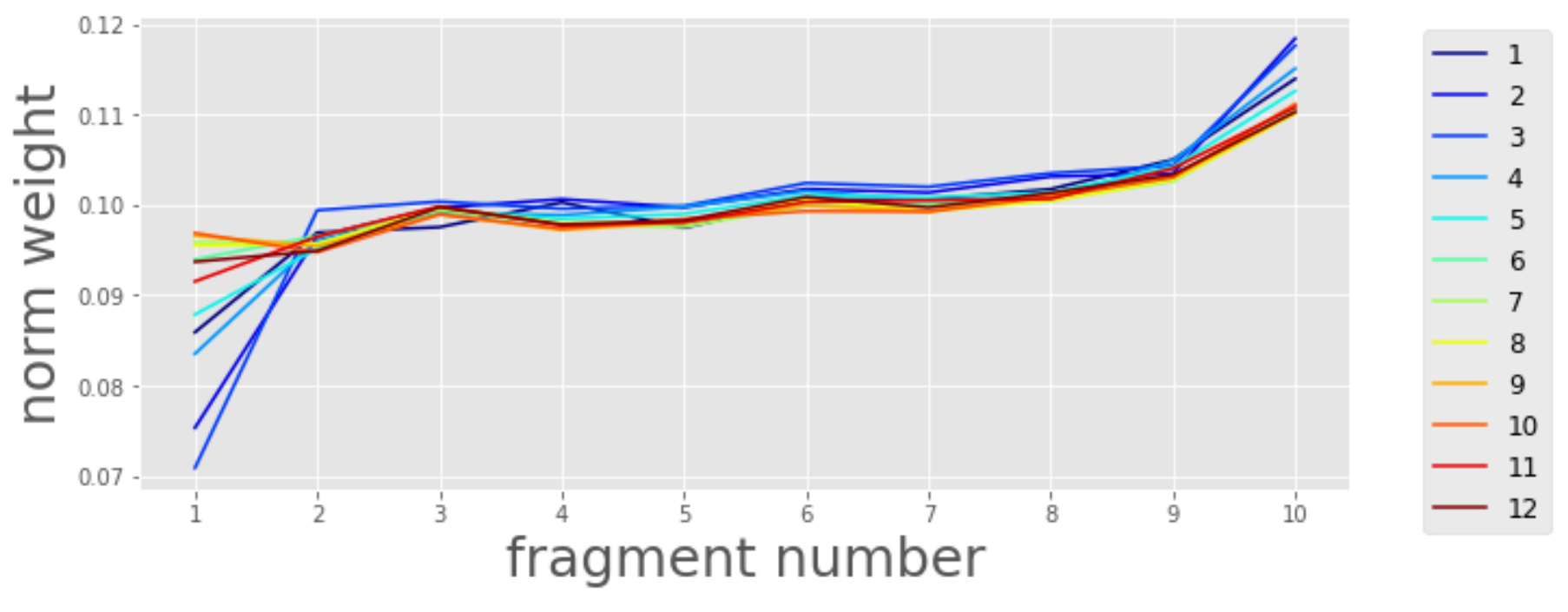}
  \vspace{-10pt}
  \caption{
Norm weights of the SID task for each fragment. 
The $l$-th line represents using the representations extracted from the $l$-th layer of HuBERT-Base to train the SID task.
There are 12 lines corresponding to the 12 layers in HuBERT-Base. 
}
  \vspace{-5pt}
  \label{fig:norm_test_without_silence}
\end{figure}

\begin{figure}[t]
  \centering
  \includegraphics[width=\linewidth]{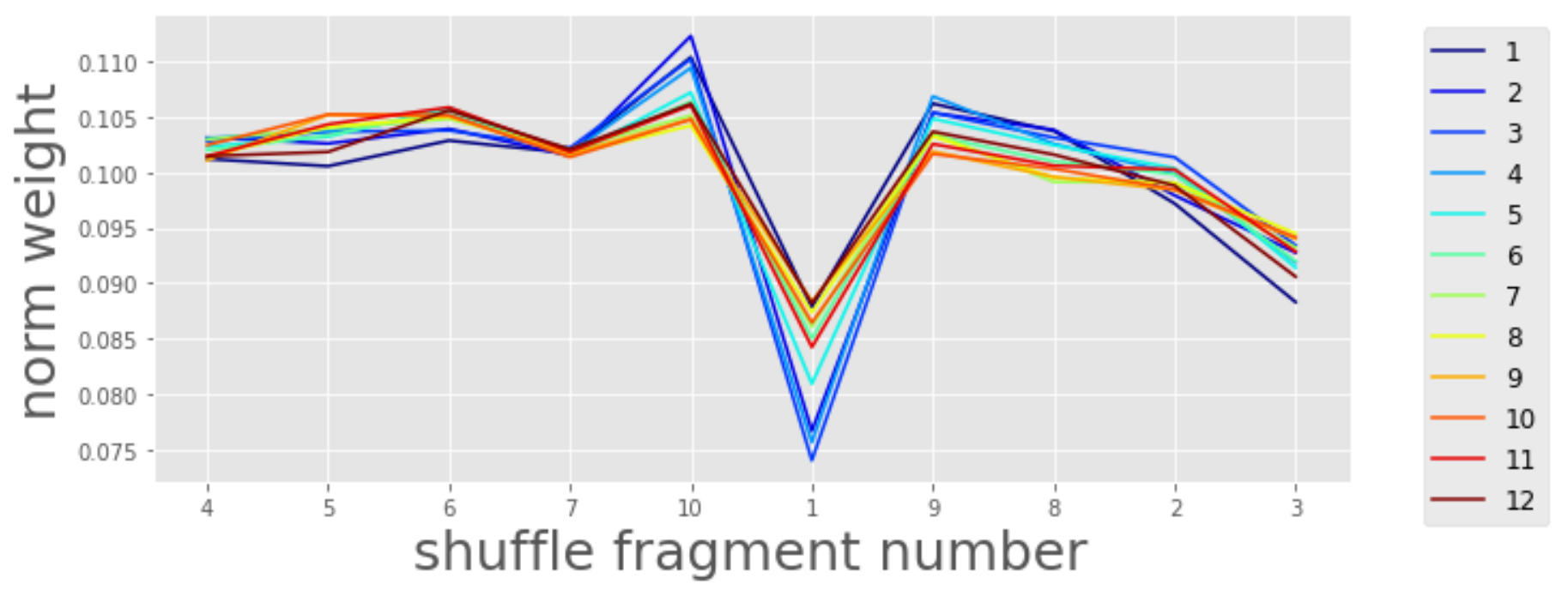}
  \vspace{-10pt}
  \caption{
Norm weights of the SID task for the shuffle experiment. 
We shuffle the waveform fragments before feeding them to HuBERT-Base. The x-axis indexes represent the original fragment numbers in Figure~\ref{fig:norm_test_without_silence}.
}
  \vspace{-5pt}
  \label{fig:shuffle_test_without_silence}
\end{figure}

\section{WHERE IS THE SPEAKER INFORMATION}
\label{sec:WHERE IS THE SPEAKER INFORMATION}




Some previous works have shown that SSL models tend to store speaker information in the middle layers~\cite{Unispeech, chen2021wavlm}.
However, how the information is stored in different positions within an utterance remains unexplored. In this section, we first analyze the importance of representations from different positions for the SID task.

The setup of the SID task here is modified from SUPERB~\cite{s3prl}. 
We follow the setting in \cite{s3prl} to split the VoxCeleb dataset~\cite{voxceleb1} into training and testing sets.
We use HuBERT as the upstream model. 
Unlike the experiments in the SUPERB benchmark, to analyze each layer, we use a single layer representation instead of the weighted sum of all layers.

In the original setting of the SID task in SUPERB, the downstream model is a pooling layer and a liner layer.  
Here we modify the pooling layer to further investigate whether SSL models store speaker information in all representations equally. The modified SID process is shown in Figure~\ref{fig:structure of experiment}(a). 
We first equally segment the sequence of representations from a specific layer in the upstream model into 10 fragments.
We do mean pooing among the frame-level representations in each fragment to obtain a representation for each fragment $f_i$ ($i=1$ to $10$).
Then we assign a learnable weight $w_i$ to the $i$-th fragment, which is learned with the downstream model.
We multiply the representation of each fragment by the learnable weight, and all the fragment representations are added to obtain a single representation $f$ for the whole utterance:
\begin{equation}
f=\sum_{i=1}^{10} w_i f_i
\end{equation}
Such a design allows us to observe whether representations in different positions of an utterance have various contributions to SID. 
If a specific fragment has a larger weight than the others, we can hypothesize that it is more critical to the SID task. 

However, it does not always mean that the corresponding fragment is critical if we only consider the absolute value of its weight. 
If a fragment's representation has small values, its weight may become large to balance the small values in the representation. 
Therefore, we consider the values of weights and representations together and introduce \textit{norm weight} $\bar{w}_i$ to indicate the importance of a fragment. 
The main idea of norm weight $\bar{w}_i$ is multiplying the learnable weight $w_i$ for each fragment with the L2 norm of the fragment's representation $\| f_i \|$. 
\begin{equation}
    \bar{w}_i  = w_i \| f_i \|
\end{equation}

Figure~\ref{fig:norm_test_without_silence} shows the average of the norm weights $\bar{w}_i$ of the fragments across the testing set.
For a reasonable comparison between utterances, we normalize the sum of all norm weights in the same utterance to 1 before taking the average. 
We find that the last fragment of the waveform always contributes the most among all the fragments. 
After investigating the property of the last fragment, we find that the last fragment contains higher portions of silence than other fragments. 

To further verify that the observation is only caused by the content rather than the order of the fragments, we randomly perturb the order of the fragments in each testing utterance and evaluate them with the same trained downstream model.
Figure~\ref{fig:shuffle_test_without_silence} shows the results. All utterances perform the same perturbation. After the perturbation, the original last fragment (fragment 10) is swapped to the middle of the utterance while still having the highest norm weight. The other perturbed fragments also remain weights close to the original values,  showing that it is not the position but the content of a segment that affects its norm weight.
Based on the finding, we hypothesize that silence fragments may be more related to the storage of speaker information.

\begin{figure}[t]
  \centering
  \includegraphics[width=\linewidth]{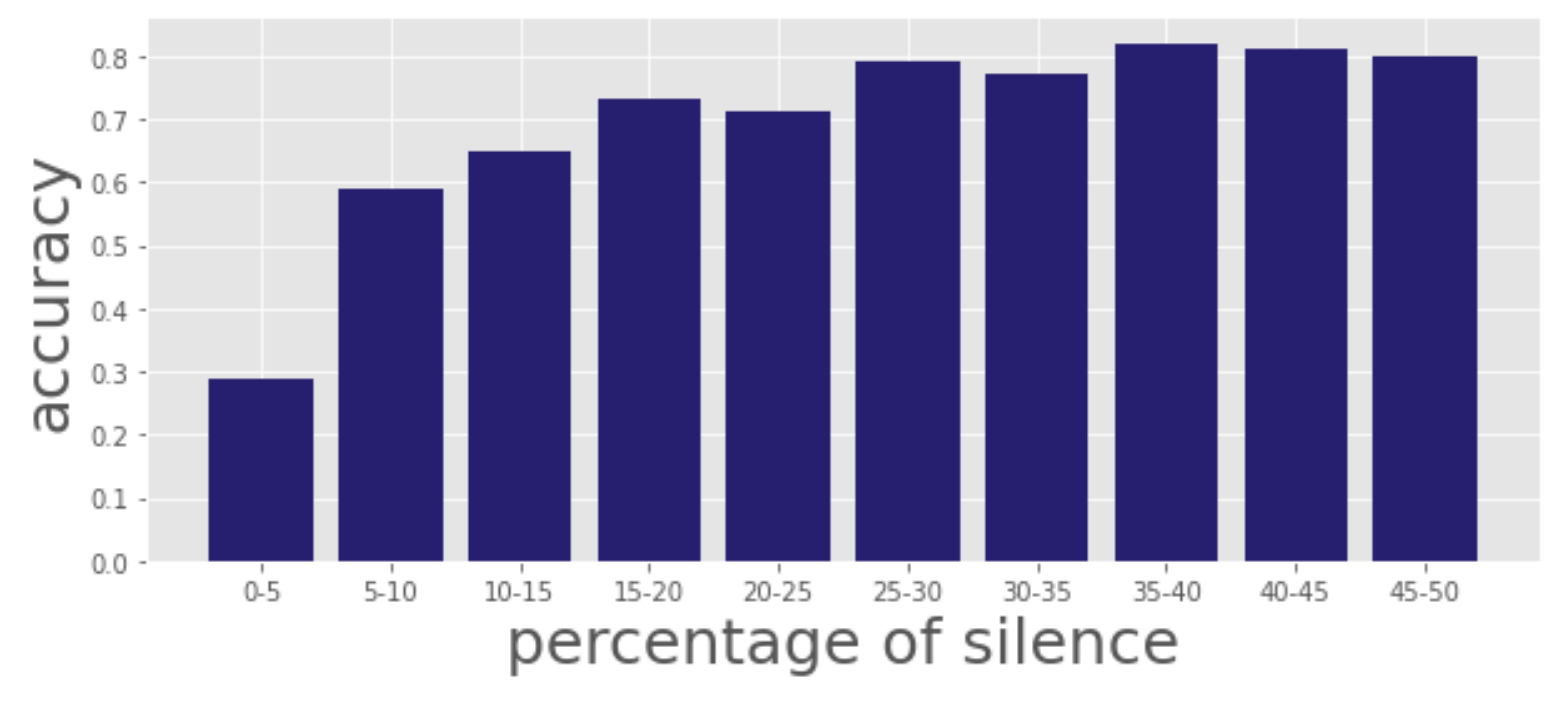}
  \vspace{-20pt}
  \caption{Relationship between the percentage of silence in waveform and the accuracy of the SID task.}
  \label{fig:relation_silence}
  \vspace{-7pt}
\end{figure}

\begin{figure}[t]
  \centering
  \includegraphics[width=\linewidth]{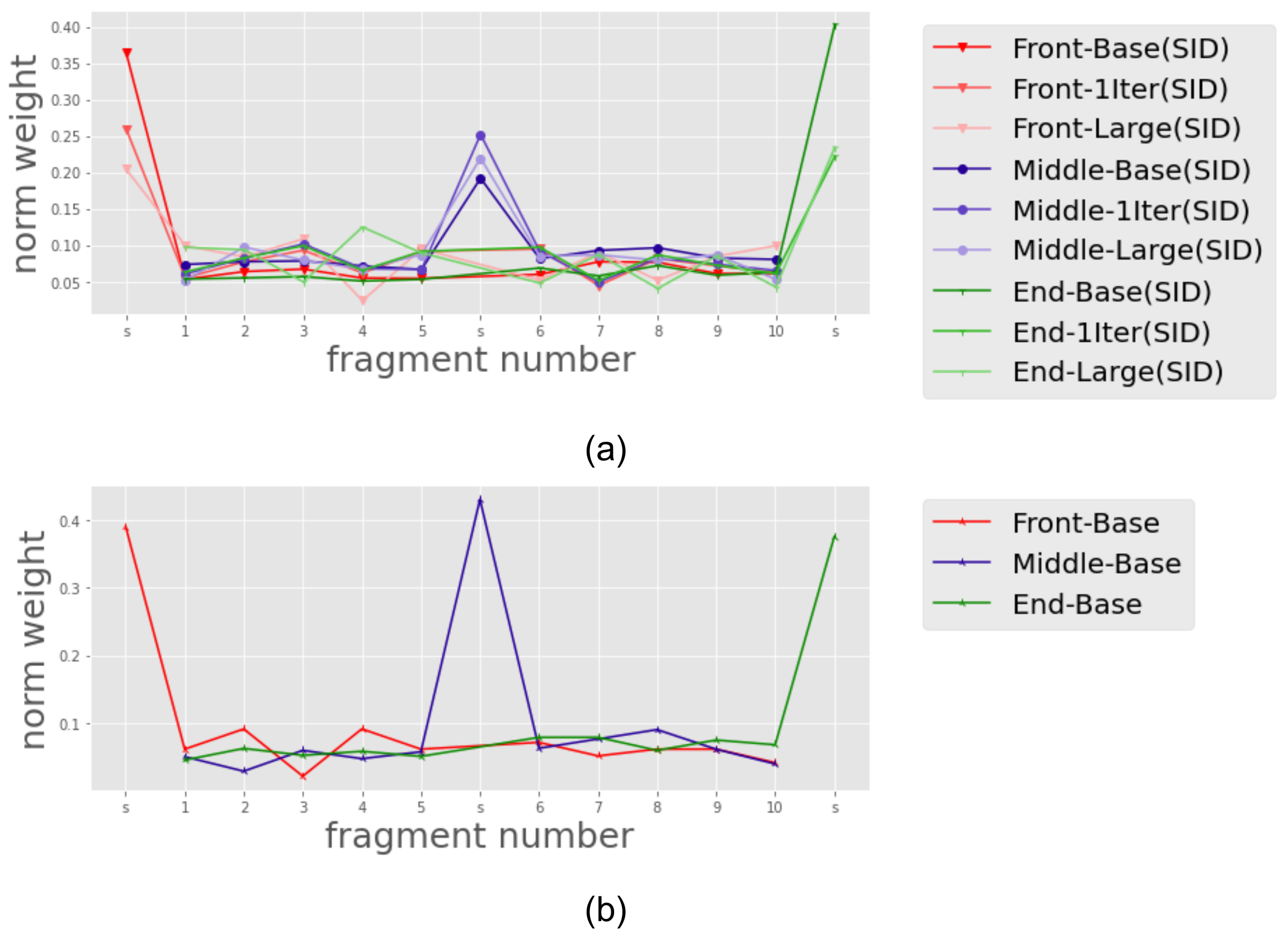}
  \vspace{-10pt}
  \caption{(a) Norm weights of the SID task with added silence. Three SSL models are used in the experiment: HuBERT-Base, HuBERT-1Iter, HuBERT-Large; (b) norm weights of the ASV task with added silence. On the x-axis, s represents the silent fragment. Red lines are silence added in the front. Blue lines are silence added in the middle. Green lines are silence added in the end.}
  \label{fig:norm_test}
  \vspace{-13pt}
\end{figure}

\begin{table}[h]
  
  \caption{SID accuracy using different fragments of HuBERT representations. The first column is the fragment number, including the silence fragment (S). The second and third columns are the silences connected at the front and end of the input waveform. The fourth column is the waveform without adding silence; the percentage in parentheses is the difference in accuracy between No Silence and Silence Front. We use McNemar's test to see if there is a significant difference between each pair of Silence Front and No Silence, and $\dag$ denotes that the results are significantly different.}  
  \vspace{-5pt}
  \label{tab:fragment_result}
  \centering
  \begin{tabular}{ c | c c c }
    \toprule
    \textbf{$F_i$} & \textbf{Silence Front} & \textbf{Silence End} & \textbf{No Silence} \\

    \midrule
    S  & \textbf{0.499} & X & X            \\
    1  & 0.337 & 0.382 & 0.418(+18\%)$^{\dag}$                \\
    2  & 0.364 & 0.364 & 0.418(+12\%)$^{\dag}$                \\
    3  & 0.374 & 0.415 & 0.440(+14\%)$^{\dag}$                  \\
    4 & 0.376 & 0.425 & 0.425(+11\%)$^{\dag}$                  \\
    5 & 0.378 & 0.425 & 0.434(+12\%)$^{\dag}$                  \\
    6 & 0.373 & 0.428 & 0.430(+13\%)$^{\dag}$                  \\
    7 & 0.380 & 0.431 & 0.439(+13\%)$^{\dag}$                  \\
    8 & 0.400 & 0.433 & 0.423(+5\%)$^{\dag}$                  \\
    9 & 0.410 & 0.430 & 0.429(+4\%)                 \\
    10 & 0.416 & 0.354 & 0.423(+1\%)                 \\
    S & X & \textbf{0.536} & X                 \\

    \bottomrule
  \end{tabular}
  \vspace{-15pt}
\end{table}

\begin{table*}[h]
  
  \caption{SID accuracy of different SSL models and silence lengths. The first column is the place where silence was added to the original waveform; the second column is the silence length compared to the original waveform. The other columns are the performance of three different models. We use McNemar's test to see if there is a significant difference between each pair of Baseline and other settings in a specific model, and $\dag$ denotes that the results are significantly different.}
  \vspace{-5pt}
  \label{tab:sid_improve}
  \centering
  \begin{tabular}{ c | c c c c }
    \toprule
    \textbf{Silence position} & \textbf{Silence length} & \textbf{HuBERT-Base} & \textbf{HuBERT-Large} & \textbf{wav2vec2}\\
    
    \midrule
    Baseline  & X & 0.807 & 0.890 & 0.739            \\
    Front  & $1/5$ & 0.803 & 0.874 & 0.735$^{\dag}$            \\
    Front  & $1/10$ & \textbf{0.824}$^{\dag}$ & \textbf{0.892}$^{\dag}$ & \textbf{0.748}$^{\dag}$            \\
    Front  & $1/20$ & 0.818$^{\dag}$ & 0.888$^{\dag}$ & 0.744            \\
    End  & $1/5$ & 0.801 & 0.878$^{\dag}$ & 0.724$^{\dag}$            \\
    End  & $1/10$ & 0.816$^{\dag}$ & 0.884 & 0.747$^{\dag}$            \\
    End  & $1/20$ & 0.813$^{\dag}$ & 0.883$^{\dag}$ & 0.746$^{\dag}$            \\
    \bottomrule
  \end{tabular}
  \vspace{-5pt}
\end{table*}

\section{WHERE CAN WE FIND SPEAKER? SILENCE}

At the end of Section~\ref{sec:WHERE IS THE SPEAKER INFORMATION}, we hypothesize that silence correlates with speaker information. 
In this section, we do more experiments to verify their relationship. 
In Section~\ref{ssub:the relationship between the amount of silence and the performance}, we show the correlation between the performance of SID with the portion of silence in an utterance.
Then in Sections~\ref{ssub:silence is important for ssl} and~\ref{ssub:do silence really important}, we insert the silence into waveforms and show the importance of the silence part to the SID task.

\subsection{Amount of Silence vs SID Task Performance}
\label{ssub:the relationship between the amount of silence and the performance}

After investigating the property of the last fragments and finding out that the last fragments contain a lot of silence, we want to know whether the amount of silence in the utterance affects the accuracy of the SID task. If the amount of silence and accuracy in the SID task are positively correlated, we can assume silence has some relationship with speaker information.

In this experiment, we use the test dataset of VoxCeleb as the measurement object. 
We use the Librosa toolkit to measure the ratio of the length of speech less than 10db to the length of the original speech in each utterance. The final result is shown in Figure~\ref{fig:relation_silence}. 
We find that if silence is less than a threshold, about 5\%, performance will be reduced by about 30\% to 50\% compared to other cases. 
With the results of this experiment, we observe that silence strongly correlates with the accuracy of the SID task in this experiment. This means that the silent part of the waveform is indeed related to the speaker's information.

\subsection{Silence is Important For SSL Model to Store Speaker Information}
\label{ssub:silence is important for ssl}
To measure the importance of silence, we add a silence fragment to the waveform, which length is 1/10 of the original waveform. The positions where we insert the silence fragment into the waveform are the original waveform's front/middle/end, respectively. After that, following the analysis method in Section~\ref{sec:WHERE IS THE SPEAKER INFORMATION}, we segment a representation sequence into 11 fragments (10 corresponding to the original waveform and 1 corresponding to the silence part we insert) and do the same analysis as in Section~\ref{sec:WHERE IS THE SPEAKER INFORMATION}. We hope to confirm whether the silence fragments contribute more to the SID task by observing the norm weights. The dataset we use is also the test dataset of VoxCeleb as in Section~\ref{sec:WHERE IS THE SPEAKER INFORMATION}. Unlike the previous experiments, to confirm that our results are general, we use three models in this section: HuBERT-Base, HuBERT-1Iter (HuBERT only trains for one iteration), and  HuBERT-Large. Furthermore, we use the weighted sum of the outputs of each layer as the representation of these three models.

The outcome is shown in Figure~\ref{fig:norm_test}. We find that no matter where we insert silence in the front/middle/end of the original waveform, the norm weight of the silence fragment is always the largest, which means that the downstream model will mainly use this fragment to classify the speakers. 
We also do the same analysis on Automatic Speaker Verification (ASV).
The setup of the ASV task here is the same as in SUPERB, except that the pooling layer is modified as in Section~\ref{sec:WHERE IS THE SPEAKER INFORMATION}.
Besides, this phenomenon also occurs in the ASV probing task, which is also shown in Figure~\ref{fig:norm_test}. 
Figure~\ref{fig:norm_test} shows that the representations corresponding to silence fragments store more speaker information than other representations.

\subsection{Do Silence Really Important for SSL Models? Yes}
\label{ssub:do silence really important}
In Section~\ref{ssub:silence is important for ssl}, the norm weights show that the silence fragments are essential for speaker information. 
This section provides more experiments to verify the silence fragments store more speaker information than the others.
This subsection also separates a representation sequence into multiple fragments, as in previous experiments. 
Still, unlike the earlier experiments, we only pick one of the fragments as the input of the SID downstream model. Besides, we pick HuBERT as our upstream model and use the weighted sum of the outputs of each layer as the representation.
The process of the experiments in this subsection is shown in Figure~\ref{fig:structure of experiment}(b).

The results are shown on Table~\ref{tab:fragment_result}.
In the columns labeled with "Silence Front" and "Silence End", we added the silence in the front or end of the original waveform, respectively.
Then we separate the representations into 11 fragments (10 corresponding to the original waveform, 1 corresponding to the silence part we insert).
Compared to other fragments, the silence fragment outperforms other fragments with about 10\% accuracy whether silence is added at the front or end. This means that the silence fragments contain more speaker information, so the SID performance is better than other fragments.

In the column labeled "No Silence",  we don't add silence in the original waveform. 
After getting the representations, we separate a representation sequence into 10 fragments and also pick one of them to train the downstream model. 
The accuracy of non-silence fragments in the waveform, which adds the silence, in the beginning, is lower than the accuracy of corresponding fragments, which don't insert the silence.
The results show that silence fragments aggregate speaker information from other fragments.  
These two experiments show that the silence fragments contain the speaker information and aggregate speaker information from other fragments.
The observations suggest that adding silence can serve as an alternative to disentangling speakers and content information.
We leave the idea as our future work.

\section{Silence Can Help to Improve the SID Task}
\label{ssub:silence can help to improve}

The previous experiments found that silence parts in waveforms will help the SSL model store speaker information. 
In this subsection, we try to utilize the above findings to improve the performance of the SID task. 
Inspired by the result of Section~\ref{ssub:do silence really important}, we directly add silence in all the data in VoxCeleb. 
The place we add is the front or end, and the length of silence is equal to 1/5, 1/10, and 1/50 of the length of the original waveform. 
We use this modified dataset to train on the SID task.

The outcome is shown in Table~\ref{tab:sid_improve}. 
We evaluated three different upstream models here: HuBERT, HuBERT-Large, wav2vec2. 
We use layer-wise weighted sum strategies in this subsection as in SUPERB, so the results of SID here are much better than the previous experiments and comparable with the results of the SUPERB benchmark.  
We find out that by using this native strategy to modify the original dataset, SID accuracy increases about 2\% for HuBERT upstream model, and accuracy increases about 0.2\% for HuBERT-Large.
The results show that adding silence into waveforms can efficiently help the model learn speaker information. Furthermore, We found that adding too much silence or too little silence is not a good thing. All three models perform best when the length of silence is 1/10. This may be because the utterance itself contains some silence, so adding too much silence will make the model disperse speaker information. Besides, HuBERT Large may be unable to make significant progress because its original performance is already as high as 90\%, so the phenomenon that adding silence still can help HuBERT-Large to improve in the SID task represent silence can help store speaker information.

\section{CONCLUSION}
In this work, we utilize SID probing task and HuBERT model to
understand how the SSL model processes and stores the speaker
information in the waveform. We use Upstream-Downstream
model structure to do probing tasks to understand
speaker information. We fix HuBERT model as our upstream
model and use a simple linear layer as our downstream model.
During the experiments, we find out that speaker information is
stored in the silence fragment. Besides, the data shows that when
the ratio of silence in the waveform is lower than 5\%, the accuracy
of the SID task will decrease by about 30\% to 50\%. Last but not
least, when adding some silence in the original waveform, the SID
task can increase up to 2\% without fine-tuning the upstream
model. These findings help us to gain more insight into how
SSL models process speaker information and can help others
to have another idea when dealing with speaker-related tasks.
In future directions, we will investigate whether different information is also stored in a specific place in the representations.
For example, whether content information is mainly stored in
the sound fragments. Moreover, we will also examine more
SSL with different structures and criteria.

\section{ACKNOWLEDGEMENTS}
We are grateful to Abdelrahman Mohamed for his comments and discussions of this paper.

\clearpage
\bibliographystyle{IEEEtran}
\bibliography{interspeech}

\end{document}